\documentclass{article}

\usepackage[preprint]{neurips_2026}

\usepackage{amsmath}
\usepackage{graphicx}

\usepackage[utf8]{inputenc} 
\usepackage[T1]{fontenc}    
\usepackage{hyperref}       
\usepackage{url}            
\usepackage{booktabs}       
\usepackage{amsfonts}       
\usepackage{nicefrac}       
\usepackage{microtype}      
\usepackage{xcolor}         

\title{VGGT-Edit: Feed-forward Native 3D Scene Editing with Residual Field Prediction}

\author{
Kaixin Zhu$^{1}$\thanks{Equal Contribution.}, 
Yiwen Tang$^{2}$*\thanks{Project Lead}, 
Yifan Yang$^{1}$*, 
Renrui Zhang$^{3}$*$\dagger$, 
Bohan Zeng$^{1}$,\\ 
Ziyu Guo$^{3}$, 
Ruichuan An$^{1}$,
Zhou Liu$^{1}$,
Qizhi Chen$^{4}$,
Delin Qu$^{4}$, 
Jaehong Yoon$^{5}$, 
Wentao Zhang$^{1,6,7}$\thanks{Corresponding Author.}
\vspace{0.2cm}\\
\textsuperscript{\rm 1}Peking University 
\textsuperscript{\rm 2}Tencent 
\textsuperscript{\rm 3}The Chinese University of Hong Kong
\textsuperscript{\rm 4}Shanghai AI Lab\\
\textsuperscript{\rm 5}NTU Singapore  
\textsuperscript{\rm 6}Zhongguancun Academy 
\textsuperscript{\rm 7}Beijing Key Lab of Data Intel. \& Security (PKU)
}

\begin{document}

\maketitle

\begin{abstract}
High-quality 3D scene reconstruction has recently advanced toward generalizable feed-forward architectures, enabling the generation of complex environments in a single forward pass. However, despite their strong performance in static scene perception, these models remain limited in responding to dynamic human instructions, which restricts their use in interactive applications. Existing editing methods typically rely on a “2D-lifting” strategy, where individual views are edited independently and then lifted back into 3D space. This indirect pipeline often leads to blurry textures and inconsistent geometry, as 2D editors lack the spatial awareness required to preserve structure across viewpoints. To address these limitations, we propose VGGT-Edit, a feed-forward framework for text-conditioned native 3D scene editing. VGGT-Edit introduces depth-synchronized text injection to align semantic guidance with the backbone's spatial poses, ensuring stable instruction grounding. This semantic signal is then processed by a residual transformation head, which directly predicts 3D geometric displacements to deform the scene while preserving background stability. To ensure high-fidelity results, we supervise the framework with a multi-term objective function that enforces geometric accuracy and cross-view consistency. We also construct the DeltaScene Dataset, a large-scale dataset generated through an automated pipeline with 3D agreement filtering to ensure ground-truth quality. Experiments show that VGGT-Edit substantially outperforms 2D-lifting baselines, producing sharper object details, stronger multi-view consistency, and near-instant inference speed. The project page is \url{https://chriszkxxx.github.io/VGGT-Edit/}.
  
\end{abstract}

\begin{figure}[ht]
    \centering
    \includegraphics[width=0.62\linewidth]{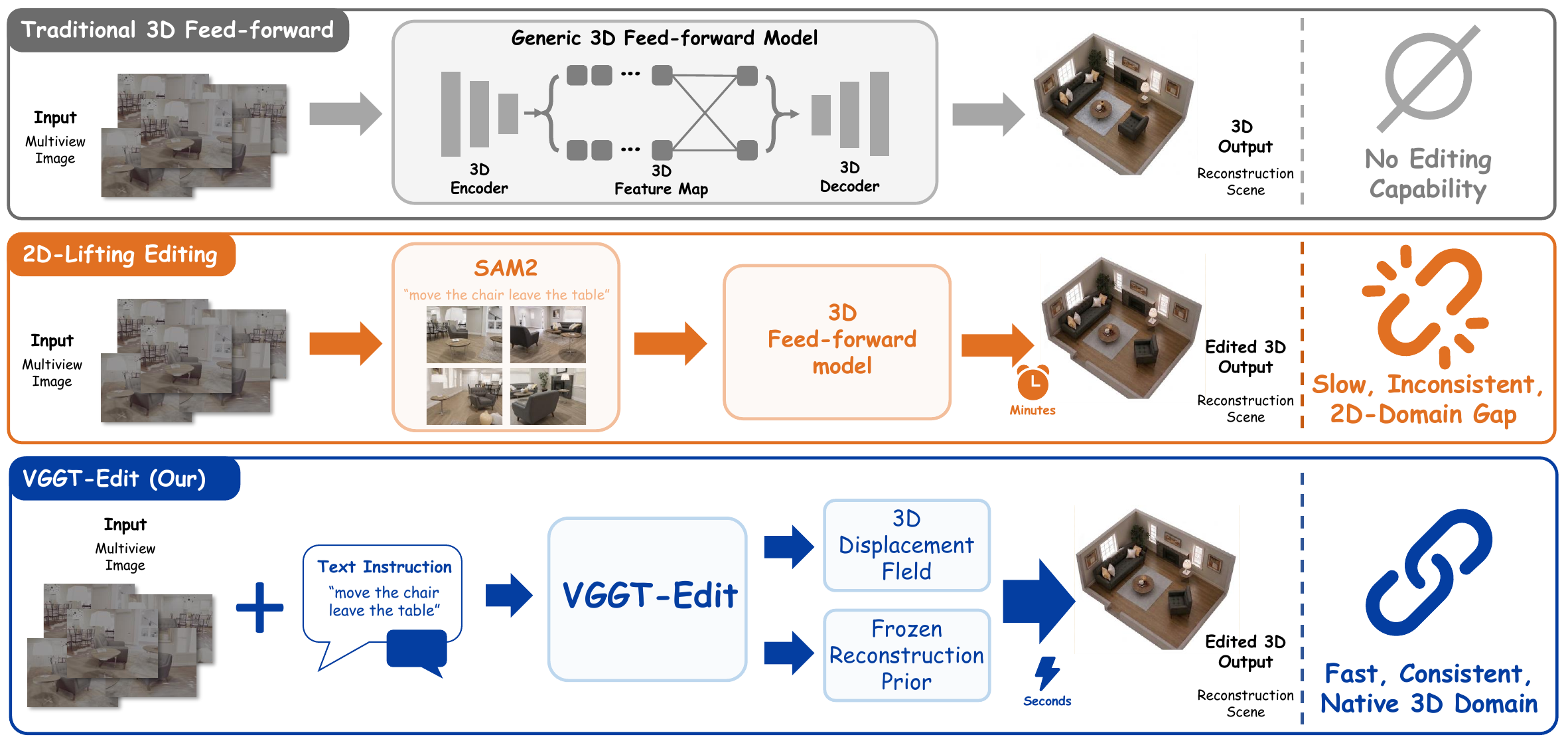}
    \caption{Comparison of 3D Editing Methods.}
    \label{fig:teaser}
\end{figure}

\section{Introduction}
High-quality 3D scene reconstruction and understanding are essential for autonomous systems and spatial computing~\cite{deitke2023objaverse,fan2024large, szymanowicz2024splatter, ravi2025sam, tang2025we, yang2025widerange4d, zeng2024trans4d,tang2024point}. Recently, the field has shifted from time-consuming per-scene optimization to generalizable feed-forward architectures~\cite{team2026openworldlib, zeng2026research,guo2023viewrefer}. Models such as VGGT~\cite{wang2025vggt} and $\pi^3$~\cite{wang2025pi} represent this emerging paradigm, enabling complex 3D environments to be reconstructed from sparse input images in a single forward pass. By avoiding expensive iterative optimization for each new scene, these methods provide an efficient geometric foundation for real-time applications.

However, fast reconstruction does not naturally imply editable scene understanding ~\cite{tang2024lgm, you2026instainpaint, wang2024dust3r}. Existing feed-forward models~\cite{yuan2026infinitevggt,wang2025vggt,maggio2025vggt} are mainly designed for static perception and lack mechanisms to respond to dynamic human instructions~\cite{chen2024mvsplat360}. Current 3D editing methods~\cite{wu2024gaussctrl,chen2024gaussianeditor,liu2025edit3r,liyi2026omni} often rely on a “2D-lifting” pipeline, where individual views are edited independently using 2D image editors and then input into the reconstruction model~\cite{sella2023vox, li2024focaldreamer, wang2024view, chen2024shap, hyung2023local,tang2024any2point}. This indirect process is inherently limited for complex scene transformations: because different views are processed separately, it often breaks multi-view geometric consistency and fails to produce stable 3D structures. This limitation is especially problematic for high-precision applications such as robotic manipulation and interactive simulation, where 3D control is required.

To address these challenges, we propose VGGT-Edit, a feed-forward framework for text-conditioned native 3D scene editing. Unlike optimization-based editing methods, VGGT-Edit performs complex scene modifications in a single forward pass. Built on a pre-trained reconstruction backbone, our method introduces a lightweight residual field prediction paradigm. Instead of re-learning the entire scene, VGGT-Edit treats editing as an incremental update to a strong geometric prior and focuses on predicting precise geometric displacements in the 3D field. This design preserves background structure while enabling controllable local modifications.

VGGT-Edit is built upon three synergistic technical components. First, we design a multimodal prompt injection module that maps linguistic intent directly into the 3D geometric space. Specifically, depth-synchronized attention aligns instruction embeddings with the backbone’s intrinsic pose-modulated features, enabling semantic guidance to be fused at the same feature depth where spatial geometry is represented. In addition, a view-aware weighting mechanism dynamically prioritizes viewpoints with clearer observations, reducing noise and artifacts caused by occlusions and camera boundaries. Second, a dedicated residual transformation head predicts a dense displacement field from the spatially fused features. By adding the predicted residuals directly to the base geometry, the model preserves the structural integrity of unchanged regions while concentrating its capacity on the target edit. Third, we introduce a residual-oriented training objective, including masked scale alignment to address global reconstruction ambiguity, as well as normal and projective consistency losses to enforce fine-grained geometry and multi-view alignment. Together, these components enable VGGT-Edit to perform precise scene transformations, such as moving or resizing objects, while maintaining efficient feed-forward inference.

To evaluate the effectiveness of our framework, we conduct extensive experiments on the DeltaScene test set, which consists of 500 high-quality and diverse 3D editing cases. Quantitative results demonstrate that VGGT-Edit significantly outperforms state-of-the-art baselines. Specifically, our model achieves a 30.2 CLIP~\cite{radford2021learning} Score, representing a 1.3 point improvement over the best existing method, while reducing the C-FID to a record low of 122.4. Most notably, our framework reduces the per-scene editing time to approximately 5 seconds. This performance represents a speedup of about 2 to 120 times compared to current 2D-lifting and optimization-based approaches. These results confirm that VGGT-Edit provides a practical and efficient foundation for interactive 3D scene manipulation. Our contributions are as follows:

\begin{itemize}
    \item \textbf{VGGT-Edit Framework}: We propose a native feed-forward 3D scene editing framework that operates directly in the geometric field, eliminating the multi-view inconsistency and high latency inherent in traditional 2D-lifting approaches.
    \item \textbf{Synchronized and Weighted Fusion Mechanism}: We design a depth-synchronized feature injection strategy together with a view-aware weighting mechanism, enabling stable, controllable, and instruction-driven 3D editing.
    \item \textbf{DeltaScene Dataset and Automated Pipeline}: We develop a scalable data generation pipeline featuring 3D agreement filtering to construct the DeltaScene dataset. This large-scale dataset provides approximately 100,000 high-quality training pairs.
    \item \textbf{Superior Performance}: Our method achieves state-of-the-art results in both geometric accuracy and multi-view consistency. Furthermore, VGGT-Edit enables near-instantaneous inference, providing a practical and efficient foundation for interactive applications in spatial computing and robotics.
\end{itemize}

\section{Related Work}
\subsection{Feed-forward 3D Reconstruction}
The field of neural 3D representations has evolved rapidly since the introduction of NeRF~\cite{mildenhall2021nerf}, which initially relied on time-consuming per-scene optimization. To enhance efficiency, generalizable methods such as PixelNeRF~\cite{yu2021pixelnerf} and MVSNeRF~\cite{chen2021mvsnerf} introduced feed-forward mechanisms capable of inferring volumetric fields from sparse inputs in a single pass. Recently, the emergence of 3D Gaussian Splatting (3DGS)~\cite{kerbl20233d} has shifted the focus toward more efficient rasterization techniques, leading to the development of feed-forward Gaussian models like pixelSplat~\cite{charatan2024pixelsplat} and MVSplat~\cite{chen2024mvsplat}. Modern architectures have further pushed these boundaries by introducing pose-agnostic learning in PF3plat~\cite{hong2024pf3plat} and permutation-equivariant geometry priors in $\pi^3$~\cite{wang2025pi}, while Speed3R~\cite{ren2026speed3r} has utilized sparse attention to enable the reconstruction of large-scale environments. While these advancements provide a robust geometric foundation for passive perception, they are fundamentally designed for static recovery rather than dynamic interaction. Our work, VGGT-Edit, leverages the powerful geometric priors of $\pi^3$ to extend these feed-forward capabilities into the realm of active, instruction-conditioned scene manipulation.

\subsection{3D Scene Editing}

Traditional 3D editing frameworks, such as Instruct-NeRF2NeRF~\cite{haque2023instruct} and GaussianEditor~\cite{chen2024gaussianeditor}, primarily rely on Score Distillation Sampling (SDS) or iterative dataset updating, which often results in extreme computational latency and precludes real-time interaction. To bridge this gap, recent research has shifted toward more efficient pipelines, which can be broadly categorized into optimization-based and feed-forward 2D-lifting approaches. Methods like GaussCtrl~\cite{wu2024gaussctrl} and EditSplat~\cite{lee2025editsplat} utilize depth-conditioned diffusion to guide 3D updates, while emerging feed-forward models such as Edit3r~\cite{liu2025edit3r} and TRACE~\cite{hu2026trace} attempt to accelerate the process into a single forward pass. However, these methods remain fundamentally tied to the 2D domain, as their architectures still operate on image-space features or rely heavily on 2D-rendering consistency during both training and inference, leading to spatial ambiguities and compromised geometric integrity in complex scenarios. In contrast, VGGT-Edit introduces a native 3D residual learning paradigm that operates directly within the 3D geometric field by predicting point-level displacements on a fixed prior. By shifting from 2D-dependent modifications to native 3D residual learning, our model effectively handles sophisticated compositional operations, such as moving and deleting multiple objects simultaneously, while ensuring strict geometric stability and near-instantaneous inference.

\begin{figure}[htbp]
  \centering
  \vspace{-0.2cm}
  \centerline{\includegraphics[width=0.9\linewidth]{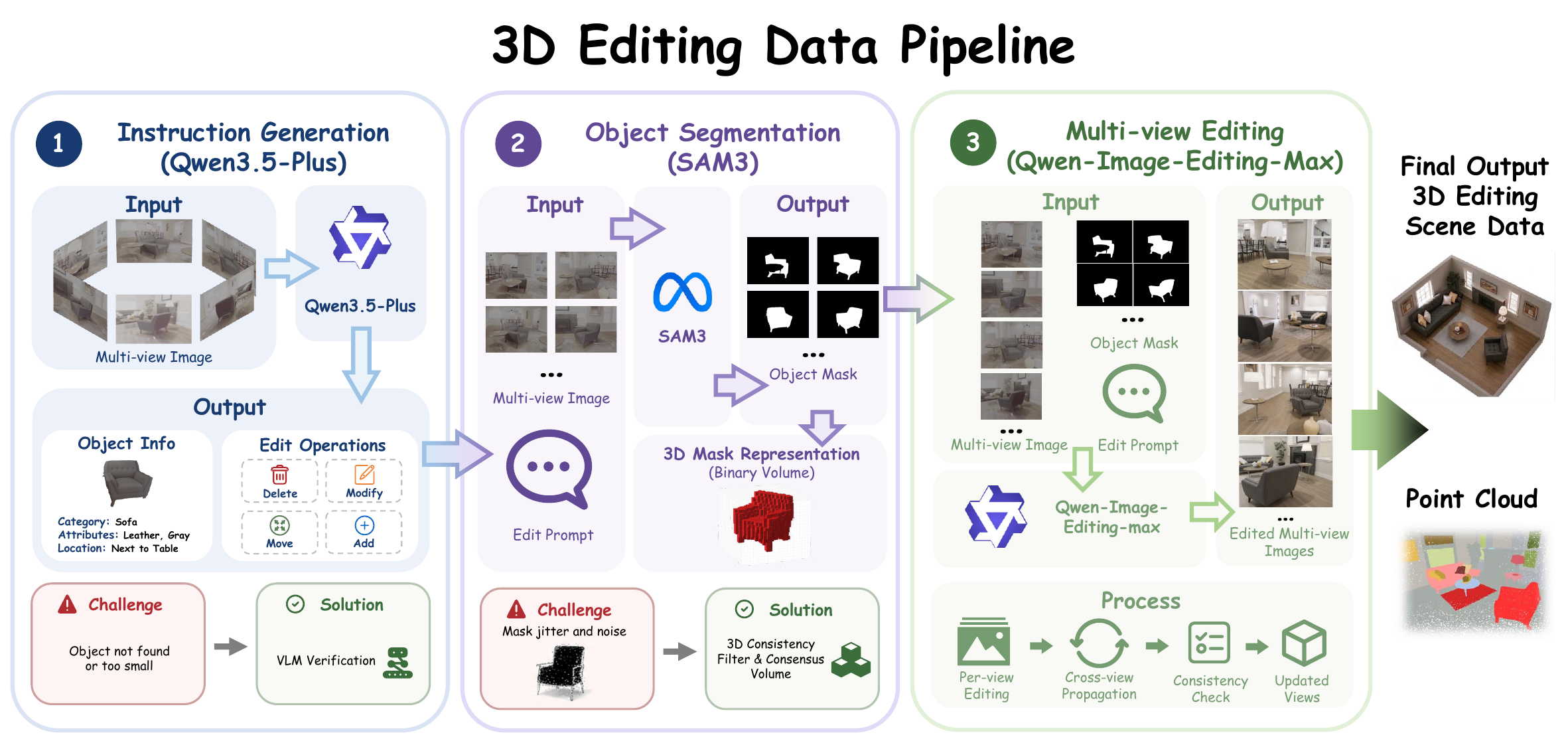}}
  \vspace{-0.2cm}
  \caption{The DeltaScene Data Pipeline. Our automated framework leverages LLMs and VLMs to generate diverse 3D editing pairs through a multi-stage consensus filtering process.}
  \label{fig:data_pipeline}
\end{figure}

\section{3D Editing Data Pipeline}

To train VGGT-Edit, we construct an automated 3D editing data generation pipeline that produces large-scale pairs of original and edited 3D scenes. Given raw multi-view observations, the pipeline converts them into high-quality, instruction-aligned, and view-consistent 3D editing pairs through four key stages. The overall design is illustrated in Fig.~\ref{fig:data_pipeline}.

\subsection{Instruction Generation and Target Selection}

The pipeline begins by using Qwen3.5-Plus~\cite{yang2025qwen3} to analyze the multi-view observations of a scene and generate candidate editing instructions. A common failure mode is that the language model may propose targets that are absent, ambiguous, or too small to support reliable 3D editing. To mitigate this issue, we introduce a VLM-based~\cite{yang2025qwen3,tang2025revision,tang2025exploring} verification step. Specifically, the LLM first proposes a set of candidate objects, and the VLM~\cite{bai2025qwen3} then verifies their visibility and spatial consistency across multiple views. Only objects that can be clearly identified in most views are retained. This process ensures that the final instruction $\mathcal{I}$ is grounded in real, visible, and geometrically valid scene content.

\subsection{3D Mask Refinement}

After selecting the target object, we use SAM3~\cite{carion2025sam} to obtain object masks in each view. However, independently predicted 2D masks often suffer from boundary noise, partial occlusions, and cross-view jitter, which can lead to inconsistent 3D supervision. To improve mask reliability, we apply a 3D consensus filtering strategy. Specifically, we project all 2D masks into 3D space and estimate a consensus volume $\bar{\mathcal{V}}$, representing the region where most views agree on the target object's location. This consensus volume is then re-projected back to each image plane to obtain a refined mask $\hat{M}_v$. A view-specific mask is considered valid only when it sufficiently overlaps with its consensus projection:
\begin{equation}
\text{Valid}(M_v) =
\begin{cases}
1, & \text{if } \text{IoU}(M_v, \hat{M}_v) > \tau, \\
0, & \text{otherwise}.
\end{cases}
\end{equation}
This refinement step reduces noisy supervision and enforces stronger multi-view consistency.

\subsection{Sequential Multi-View Editing}

A central challenge in data generation is maintaining appearance and geometry consistency of the edited target across viewpoints. Editing each view independently can introduce inconsistent colors, textures, shapes, or spatial layouts, making the resulting data unsuitable for learning native 3D scene editing. To address this issue, we adopt a sequential multi-view editing strategy. Instead of editing all views independently, we edit them in an ordered sequence and condition the current edit on the previously edited view\cite{wu2025qwen}:
\begin{equation}
\hat{I}_v = \text{Edit}(I_v, M_v, \mathcal{I}, \text{Cond}=\hat{I}_{v-1}).
\end{equation}
By propagating visual context across adjacent views, this strategy encourages consistent object appearance and spatial placement throughout the sequence. As a result, the generated editing pairs provide more reliable supervision for learning residual field prediction in 3D space.

\subsection{Viewpoint Selection and Quality Control}

Not all views provide equally reliable supervision. Some viewpoints may contain severe occlusion, truncation, extreme viewing angles, or weak target visibility. To select high-quality observations, we introduce a Re-projection Fidelity score to evaluate each view. For a given view $v$, we project its mask into 3D and then re-project it back to the image plane, obtaining a reconstructed mask $\tilde{M}_v$. The score is defined as:
\begin{equation}
RF(M_v) = \text{IoU}(M_v, \tilde{M}_v) \cdot \cos(\theta_v),
\end{equation}
where $\theta_v$ denotes the viewing angle. This metric favors views with accurate geometric projection and frontal, unobstructed observations. By filtering out unreliable views, the pipeline provides cleaner supervision and improves the stability of VGGT-Edit training.

\section{The DeltaScene Dataset}

\begin{figure}[htbp]
\centering
\vspace{-0.2cm}
\centerline{\includegraphics[width=1\linewidth]{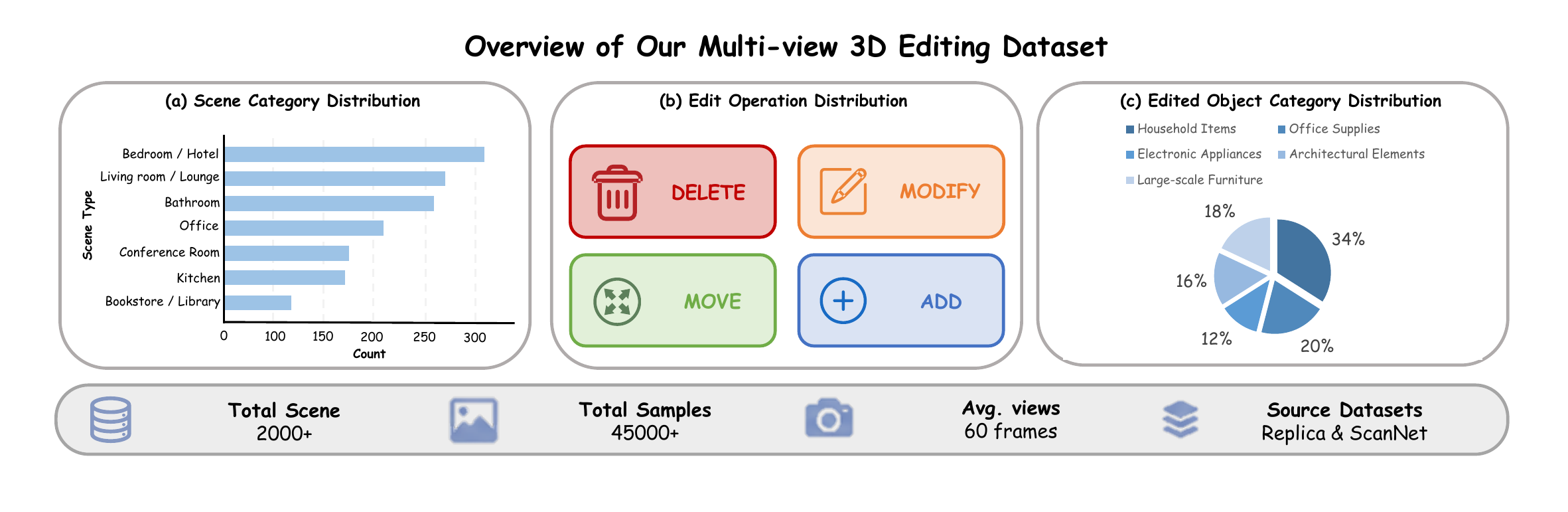}}
\vspace{-0.2cm}
\caption{Overview of the DeltaScene Dataset Details.}
\label{fig:dataset}
\end{figure}

In this section, we provide a detailed description of the DeltaScene Dataset, which is specifically constructed to address the lack of large-scale, view-consistent data for native 3D scene editing. High-quality data is fundamental to training our residual learning paradigm, as it requires precise geometric alignment between the original and edited scenes, as illustrated in Fig.~\ref{fig:dataset}.

\paragraph{Data Generation Pipeline.}
We develop an automated pipeline to generate large-scale, multi-view consistent editing pairs. The process begins with a diverse collection of high-quality 3D scene priors from sources including Replica~\cite{straub2019replica}, ScanNet~\cite{dai2017scannet}, and ScanNet++~\cite{yeshwanth2023scannet++}. For each scene, we leverage Large Language Models (LLMs) to brainstorm realistic and complex editing instructions. To ensure these edits are spatially grounded, we use Vision-Language Models (VLMs) to identify the target regions within the 3D field. We then apply a multi-view rendering engine to generate the corresponding "before" and "after" image sequences. All editing pairs are refined using 3D consensus filtering and re-projection fidelity scoring, ensuring every edit maintains strict geometric consensus across all viewpoints and providing the necessary ground truth for residual displacement learning.

\paragraph{Dataset Statistics and Diversity.}
DeltaScene consists of approximately 100,000 high-quality editing pairs (including 95,000 training and 500 manual-verified testing samples), covering a wide range of indoor and outdoor environments such as offices, living rooms, and residential spaces. The dataset is designed to be operationally and semantically diverse, incorporating four atomic 3D editing operations: (1) \textbf{Add}, which inserts new style-matched elements; (2) \textbf{Delete}, which removes target objects and recovers the background; (3) \textbf{Modify}, which changes object attributes such as color, material, or texture; and (4) \textbf{Move}, which alters object position or orientation. These operations further support compositional editing, where multiple modifications are applied within the same scene. To evaluate varied geometric properties, we curate a wide selection of household items, office supplies, electronic appliances, and large-scale furniture, alongside architectural elements like windows and doors. This holistic design ensures that VGGT-Edit learns a robust mapping from text instructions to complex 3D changes across any scene context.

\paragraph{Quality Control Mechanisms.}
To guarantee the reliability of our benchmarks, the 500 testing pairs underwent rigorous manual verification and refinement. Each sample was checked for both semantic accuracy—ensuring the visual change strictly follows the text instruction—and geometric stability, confirming that non-edited background regions remain perfectly static. This careful selection process, combined with our re-projection fidelity scoring, ensures that our quantitative evaluations, such as the C-FID and CLIP Score, accurately reflect the model's performance in real-world scenarios.

\section{Model}

\subsection{Overview}

\begin{figure}[htbp]
  \centering
  \vspace{-0.2cm}
  \centerline{\includegraphics[width=0.9\linewidth]{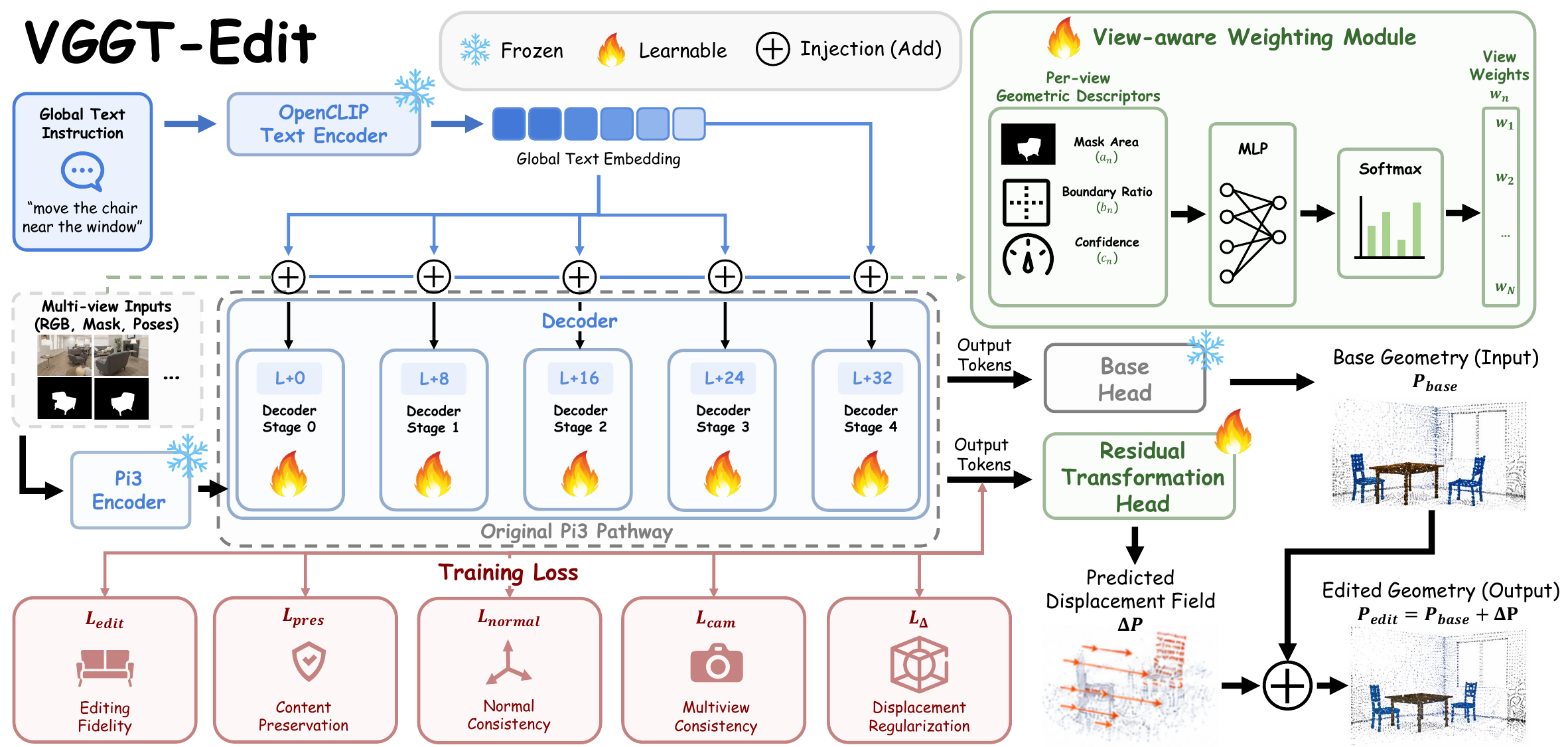}}
  \vspace{-0.2cm}
  \caption{Overview of the VGGT-Edit Model Architecture. Our model features a synchronized prompt injection mechanism and a residual transformation head, enabling native 3D scene editing through a single forward pass.}
  \label{fig:model_architecture}
\end{figure}

VGGT-Edit is designed for efficient, instruction-driven native 3D scene editing. Given sparse-view images $\{I_n\}_{n=1}^{N}$, camera parameters $\{\boldsymbol{\theta}_n\}_{n=1}^{N}$, and a text instruction $\mathcal{I}$, the model predicts an edited 3D geometry in a single forward pass. Unlike 2D-lifting methods that edit each image independently and then reconstruct the edited scene, VGGT-Edit performs editing directly in the 3D geometric field. This design avoids cross-view conflicts and enables stable, localized scene modifications.

As shown in Fig.~\ref{fig:model_architecture}, VGGT-Edit consists of three main architectural components. First, a frozen feed-forward reconstruction backbone provides a strong geometric prior. Second, a depth-synchronized text injection module aligns the editing instruction with spatially grounded multi-view features. Third, a residual transformation head predicts a dense residual displacement field, which is added to the base geometry under the guidance of an edit mask. To train this architecture, we further introduce a residual-oriented objective that combines edit reconstruction, non-edit preservation, normal consistency, camera-frame consistency, and residual regularization. This formulation enables VGGT-Edit to preserve unchanged regions and perform localized geometry deformation.

\subsection{Feed-forward Geometric Prior}

We build VGGT-Edit upon $\pi^3$~\cite{wang2025pi}, a generalizable feed-forward reconstruction model. Given $N$ sparse-view images and their corresponding camera parameters, $\pi^3$ extracts multi-view features using an image encoder $\Phi$ and a permutation-equivariant transformer $\Psi$:
\begin{equation}
\mathbf{F} =
\Psi\left(
\Phi(I_1,\ldots,I_N),
\boldsymbol{\theta}_1,\ldots,\boldsymbol{\theta}_N
\right).
\end{equation}

The backbone further predicts a dense base point map
$\mathbf{P}_{\mathrm{base}} \in \mathbb{R}^{N \times H \times W \times 3}$,
which represents the reconstructed 3D geometry of the original scene. Rather than training a scene-specific representation from scratch, we freeze the reconstruction backbone and use it as a generalizable geometric prior. This choice is important for two reasons. First, it preserves the robust spatial structure learned from large-scale reconstruction data. Second, it allows the editing module to focus on modeling the requested change rather than re-learning the entire scene geometry.

\subsection{Depth-Synchronized Text Injection}

To perform instruction-driven editing, VGGT-Edit must map the semantic intent of a text instruction to the correct spatial region in the 3D field. We therefore introduce a depth-synchronized text injection module, which injects textual guidance into the reconstruction features at layers aligned with the backbone's pose-modulation stages.

Given an instruction $\mathcal{I}$, we obtain a text embedding $\mathbf{e}_{\mathrm{text}}$ using a pre-trained OpenCLIP~\cite{ilharco2021openclip} encoder. Instead of injecting this embedding only once, we fuse it into the transformer decoder at multiple synchronized layers:
\begin{equation}
\mathcal{L} = \{l \cdot k + 1\}_{l=0}^{4}, \quad k=8.
\end{equation}
These layers are selected to match the major pose-injection blocks of the reconstruction backbone. As a result, semantic guidance is introduced at the same feature depths where spatial geometry is progressively formed.

At each selected layer, we perform text-driven cross-attention between the multi-view features and the instruction embedding. This synchronized design provides continuous semantic guidance throughout the decoding process. Compared with a single early injection, it reduces the risk that textual information fades in deeper layers. Compared with injecting text into every layer, it avoids unnecessary computation and training instability. In practice, this enables the model to produce edits that are both semantically aligned with the instruction and spatially consistent across views.

\subsection{View-Aware Importance Weighting}

Multi-view observations are not equally informative for editing. In some views, the target object may be clearly visible, while in others it may be occluded, truncated, or close to the image boundary. Treating all views equally can therefore introduce noisy semantic guidance.

To address this issue, we introduce a view-aware importance weighting mechanism. For each view $n$, we construct a geometric descriptor
\begin{equation}
\mathbf{g}_n = [s_n, a_n, c_n],
\end{equation}
where $s_n$ denotes the visible mask area, $a_n$ denotes the boundary ratio, and $c_n$ denotes the backbone confidence score. These quantities jointly describe the reliability of the target observation in that view. A lightweight MLP predicts a normalized importance weight:
\begin{equation}
w_n =
\frac{\exp(\mathrm{MLP}(\mathbf{g}_n))}
{\sum_{j=1}^{N}\exp(\mathrm{MLP}(\mathbf{g}_j))}.
\end{equation}

The resulting weight is used to modulate the key and value features derived from the text embedding:
\begin{equation}
\mathbf{K}_n = \sqrt{w_n}\mathbf{W}_k\mathbf{e}_{\mathrm{text}},
\quad
\mathbf{V}_n = \sqrt{w_n}\mathbf{W}_v\mathbf{e}_{\mathrm{text}}.
\end{equation}
This formulation allows views with more complete observations to contribute more strongly to the editing process, while suppressing unreliable views caused by occlusion or boundary artifacts.

\subsection{Residual Field Prediction}

After text injection and view-aware weighting, the spatially fused features are passed to a residual transformation head. Unlike the original reconstruction head, which predicts the full scene geometry, our head predicts only the incremental geometric change required by the instruction.

Specifically, the head outputs a dense residual displacement field
$\boldsymbol{\Delta}\mathbf{P} \in \mathbb{R}^{N \times H \times W \times 3}$.
Given an edit mask $\mathbf{M}$, the edited point map is obtained by:
\begin{equation}
\mathbf{P}_{\mathrm{edit}}
=
\mathbf{P}_{\mathrm{base}}
+
\boldsymbol{\Delta}\mathbf{P} \odot \mathbf{M},
\end{equation}
where $\odot$ denotes element-wise multiplication. This residual formulation is central to VGGT-Edit. Since most regions in a scene remain unchanged after an edit, directly predicting the complete edited geometry is unnecessary and may destabilize training. By predicting only localized residual displacements, VGGT-Edit preserves the background structure inherited from the frozen backbone and concentrates its modeling capacity on the edited region.

\subsection{Training Objectives}

We train VGGT-Edit with a multi-term objective that supervises edited geometry, preserves unchanged regions, and enforces geometric and projective consistency.

\paragraph{Masked Scale Alignment.}
Feed-forward reconstruction models may exhibit global scale ambiguity relative to ground-truth geometry. Since our goal is to learn accurate relative editing, we compute a per-sample masked least-squares scale factor within the edit region:
\begin{equation}
s =
\frac{
\sum_{n,h,w}
(\mathbf{P}_{\mathrm{edit}}^{n,h,w} \odot \mathbf{M}^{n,h,w})
\cdot
(\mathbf{P}_{\mathrm{gt}}^{n,h,w} \odot \mathbf{M}^{n,h,w})
}{
\sum_{n,h,w}
\left\|
\mathbf{P}_{\mathrm{edit}}^{n,h,w} \odot \mathbf{M}^{n,h,w}
\right\|_2^2
+\epsilon
}.
\end{equation}
The aligned prediction is then:
\begin{equation}
\hat{\mathbf{P}} = s \cdot \mathbf{P}_{\mathrm{edit}}.
\end{equation}

For compactness, we define a masked $\ell_1$ loss as:
\begin{equation}
\mathcal{L}_{1}(\mathbf{A},\mathbf{B};\mathbf{M})
=
\frac{1}{3\sum \mathbf{M}}
\sum_{n,h,w,c}
\mathbf{M}^{n,h,w}
\left|
\mathbf{A}^{n,h,w,c}
-
\mathbf{B}^{n,h,w,c}
\right|.
\end{equation}

\paragraph{Edit Reconstruction and Preservation.}
The edit reconstruction loss supervises the edited region against the target point map:
\begin{equation}
\mathcal{L}_{\mathrm{edit}}
=
\mathcal{L}_{1}
(\hat{\mathbf{P}}, \mathbf{P}_{\mathrm{gt}}; \mathbf{M}).
\end{equation}
To preserve the unchanged scene structure, we constrain the non-edit region to remain close to the frozen base geometry:
\begin{equation}
\mathcal{L}_{\mathrm{pres}}
=
\mathcal{L}_{1}
(
\mathbf{P}_{\mathrm{edit}},
\mathrm{stopgrad}(\mathbf{P}_{\mathrm{base}});
1-\mathbf{M}
).
\end{equation}
This term discourages unnecessary deformation outside the region and improves background stability.

\paragraph{Normal Consistency.}
To encourage geometrically plausible surfaces, we compute surface normals from the dense point maps using finite differences and penalize angular deviation:
\begin{equation}
\mathcal{L}_{\mathrm{normal}}
=
\frac{1}{\sum \mathbf{M}'}
\sum
\mathbf{M}'
\left(
1-
\langle
\mathbf{n}_{\mathrm{pred}},
\mathbf{n}_{\mathrm{gt}}
\rangle
\right),
\end{equation}
where $\mathbf{M}'$ is the mask resized or cropped to match the normal grid.

\paragraph{Camera-Frame Consistency.}
To further enforce multi-view geometric alignment, we supervise the prediction in camera space. We transform the aligned prediction and ground truth using the world-to-camera pose, obtaining camera-space points $\mathbf{X}^{\mathrm{cam}}$ and $\mathbf{Y}^{\mathrm{cam}}$. We then constrain both perspective rays and log-depth:
\begin{equation}
\mathcal{L}_{\mathrm{cam}}
=
\mathcal{L}_{1}
(
\mathbf{r}(\mathbf{X}^{\mathrm{cam}}),
\mathbf{r}(\mathbf{Y}^{\mathrm{cam}});
\mathbf{M}
)
+
\mathcal{L}_{1}
(
\ell(\mathbf{X}^{\mathrm{cam}}),
\ell(\mathbf{Y}^{\mathrm{cam}});
\mathbf{M}
),
\end{equation}
where $\mathbf{r}(\mathbf{x})=(x/z,y/z)$ denotes the perspective ray and $\ell(\mathbf{x})=\log z$ denotes log-depth. This loss encourages the edited geometry to remain consistent under camera projection.

\paragraph{Residual Regularization.}
Finally, we regularize the residual displacement field to avoid unnecessarily large deformations:
\begin{equation}
\mathcal{L}_{\Delta}
=
\frac{1}{3\sum \mathbf{M}}
\sum_{n,h,w}
\mathbf{M}^{n,h,w}
\left\|
\boldsymbol{\Delta}\mathbf{P}^{n,h,w}
\right\|_2^2.
\end{equation}

The final training objective is:
\begin{equation}
\mathcal{L}_{\mathrm{total}}
=
\lambda_{\mathrm{edit}}\mathcal{L}_{\mathrm{edit}}
+
\lambda_{\mathrm{pres}}\mathcal{L}_{\mathrm{pres}}
+
\lambda_{\mathrm{normal}}\mathcal{L}_{\mathrm{normal}}
+
\lambda_{\mathrm{cam}}\mathcal{L}_{\mathrm{cam}}
+
\lambda_{\Delta}\mathcal{L}_{\Delta}.
\end{equation}

\section{Experiments}

In this section, we evaluate VGGT-Edit on the proposed DeltaScene Dataset. We conduct experiments to assess its performance in geometric accuracy, multi-view consistency, semantic alignment, and inference efficiency.

\subsection{Implementation Details}
We implement VGGT-Edit in PyTorch and train it on 8 NVIDIA A100 GPUs for 10k iterations with a batch size of 16. The model is initialized with a frozen $\pi^3$ backbone~\cite{wang2025pi}, while the decoder transformer, text injection module, and residual head are trained using Adam with a learning rate of $1\times10^{-4}$ and cosine decay. The training loss weights are set to $\lambda_{\mathrm{edit}}=1.0$, $\lambda_{\mathrm{pres}}=1.0$, $\lambda_{\mathrm{normal}}=0.1$, $\lambda_{\mathrm{cam}}=0.1$, and $\lambda_{\Delta}=0.01$. We utilize 95,000 pairs from the DeltaScene Dataset for training and 500 pairs for testing, covering diverse scene categories and editing operations.

\subsection{Baselines}
We compare VGGT-Edit with representative 3D editing methods across three major paradigms to evaluate its performance. For 2D-lifting approaches, we include GaussCtrl~\cite{wu2024gaussctrl} and Omni-3DEdit~\cite{liyi2026omni}. It is important to note that GaussCtrl is a per-scene optimization method that maintains multi-view consistency through an iterative refinement process, whereas Omni-3DEdit focuses on unified image-domain editing before 3D lifting. We also compare our model with EditSplat~\cite{lee2025editsplat}, another optimization-based baseline that achieves high-fidelity results at the cost of substantial per-scene computation time. Finally, we evaluate feed-forward editing methods, including Edit3r~\cite{liu2025edit3r} and an adapted version of NoPoSplat~\cite{ye2024no}. Since NoPoSplat was originally designed as a generalizable reconstruction model, we extend it into a 2D-lifting baseline by employing Qwen-Image-Editing-Max as a pre-processor to edit input views prior to 3D reconstruction. These comprehensive comparisons highlight the advantages of our native 3D formulation, depth-synchronized text injection, and residual field prediction over both iterative optimization-based pipelines and existing feed-forward architectures.

\subsection{Main Results}

\paragraph{Evaluation Metrics.}
We evaluate VGGT-Edit from three aspects: semantic alignment, multi-view consistency, and inference efficiency. For semantic alignment, we report the CLIP Score between rendered views of the edited scene and the input text instruction. To measure cross-view visual and geometric consistency, we compute C-FID and C-KID across rendered viewpoints. These metrics capture view-dependent artifacts such as ghosting, flickering, and inconsistent object structure, which are common in 2D-lifting pipelines. Finally, we report the average inference time per edit to measure efficiency against optimization-based and diffusion-based methods.

\begin{figure}[ht]
    \centering
    \vspace{-0.2cm}
    \includegraphics[width=0.90\linewidth]{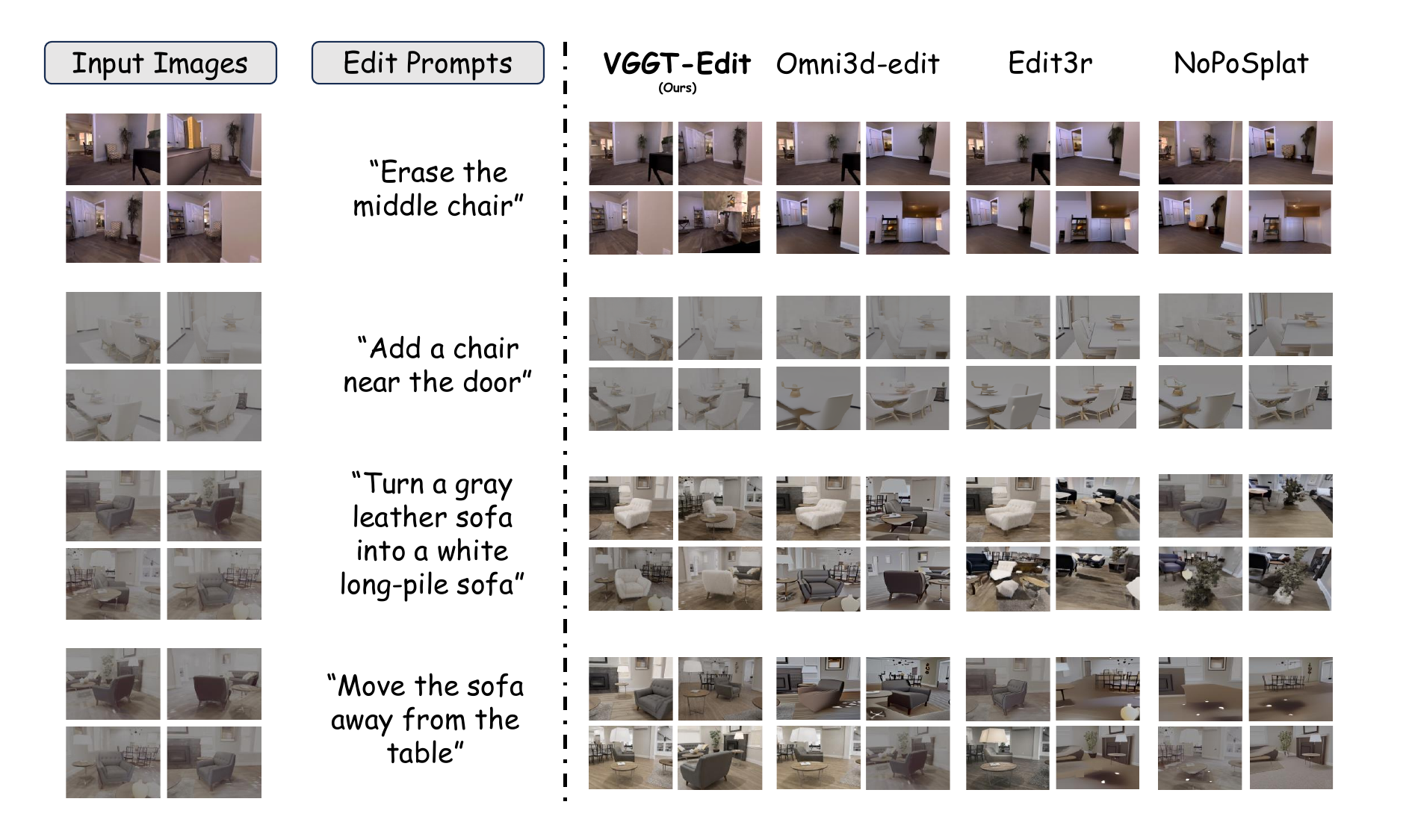}
    \vspace{-0.2cm}
    \caption{\textbf{Qualitative Comparison on Diverse 3D Editing Tasks.} We evaluate VGGT-Edit against state-of-the-art baselines across various scene-level operations, including object addition, removal, and transformation. Our method produces sharp, instruction-accurate results while maintaining superior multi-view consistency compared to 2D-lifting approaches.}
    \label{fig:main_results}
\end{figure}

\paragraph{Quantitative Comparison.}
Table~\ref{tab:main_results} summarizes the quantitative results on the DeltaScene Dataset. VGGT-Edit achieves the best overall performance across semantic alignment, multi-view consistency, and efficiency. Compared with 2D-lifting methods such as Omni-3DEdit and GaussCtrl, VGGT-Edit obtains substantially lower C-FID and C-KID scores, demonstrating that residual field prediction in the 3D geometric field better preserves cross-view structural consistency than independently editing 2D views. In addition, VGGT-Edit requires only approximately 2 seconds per scene, making it nearly two orders of magnitude faster than optimization-based methods such as EditSplat and clearly more efficient than recent feed-forward baselines such as Edit3r. These results show that VGGT-Edit provides a strong balance between high-fidelity native 3D editing and practical inference efficiency.

\begin{table}[h]
\centering
\caption{Quantitative results on the DeltaScene dataset. CLIP Score measures semantic alignment, while C-FID and C-KID evaluate geometric consistency. Time (s) denotes the end-to-end inference latency, measuring the full process from multi-view input images to the final edited 3D scene.}
\label{tab:main_results}
\begin{tabular}{l|cccc}
\hline
Method & CLIP Score $\uparrow$ & C-FID $\downarrow$ & C-KID $\downarrow$ & Time (s) $\downarrow$  \\ \hline
GaussCtrl \cite{wu2024gaussctrl} & 26.4 & 145.2 & 0.192 & $\sim$300 \\
EditSplat \cite{lee2025editsplat} & 27.1 & 138.5 & 0.154 & $\sim$600 \\
Omni-3DEdit \cite{liyi2026omni} & 28.5 & 128.1 & 0.85 & $\sim$115 \\
NoPoSplat \cite{ye2024no} & 25.8 & 135.4 & 0.112 & $\sim$20 \\
Edit3r \cite{liu2025edit3r} & 28.9 & 130.8 & 0.92 & $\sim$10 \\ \hline
\textbf{VGGT-Edit (Ours)} & \textbf{30.2} & \textbf{122.4} & \textbf{0.048} & \textbf{$\sim$5} \\ \hline
\end{tabular}
\end{table}

\paragraph{Qualitative Comparison.}
As shown in Fig.~\ref{fig:main_results}, the qualitative results further demonstrate the effectiveness of VGGT-Edit for native 3D scene editing. Existing 2D-lifting and feed-forward baselines often produce geometric misalignment, blurred boundaries, and view-dependent artifacts. For example, in the ''Erase the middle chair'' and ''Add a chair'' tasks, methods such as Edit3r and NoPoSplat exhibit ghosting effects and unstable object shapes, as their edits are not directly grounded in the 3D geometric field. In several cases, the edited regions appear as 2D overlays that are weakly attached to the underlying scene geometry.

In contrast, VGGT-Edit generates sharp and spatially grounded edits that remain consistent across camera viewpoints. By leveraging a frozen geometric prior and predicting localized residual displacement fields, our method preserves background structure while accurately executing instruction-guided modifications. These visual improvements are consistent with the quantitative results: VGGT-Edit achieves a CLIP Score of 30.2, improving the best competing method by 1.3 points, and obtains the lowest C-FID of 122.4. These results indicate that depth-synchronized text injection and residual field prediction effectively translate semantic instructions into accurate 3D deformations, leading to high visual quality and strong multi-view consistency.

\subsection{Ablation Study}

\begin{table}[h]
\centering
\caption{Ablation study of VGGT-Edit components. Sync-Attn: depth-synchronized attention; View-W: view-aware weighting; Res-Head: residual transformation head.}
\label{tab:ablation}
\begin{tabular}{ccc|ccc}
\hline
Sync-Attn & View-W & Res-Head & CLIP Score $\uparrow$ & C-FID $\downarrow$ & C-KID $\downarrow$ \\ \hline
          & \checkmark & \checkmark & 28.1 & 126.5 & 0.062 \\ 
\checkmark &           & \checkmark & 27.8 & 127.2 & 0.068 \\ 
\checkmark & \checkmark &           & 29.5 & 131.4 & 0.085 \\ 
\hline
\checkmark & \checkmark & \checkmark & \textbf{30.2} & \textbf{122.4} & \textbf{0.048} \\ \hline
\end{tabular}
\end{table}

\paragraph{Architecture Effectiveness.}
We conduct ablation studies to evaluate the contribution of each core component, with quantitative results reported in Table~\ref{tab:ablation}. Replacing depth-synchronized text injection with single-layer injection reduces the CLIP Score from 30.2 to 28.1, indicating that multi-stage semantic guidance is important for accurate instruction alignment. Removing view-aware weighting increases C-FID, suggesting that unreliable observations from occluded or boundary views introduce geometric noise. Finally, replacing the residual transformation head with a standard reconstruction head leads to the weakest geometric consistency and background stability. These results demonstrate that depth-synchronized text injection, view-aware weighting, and residual field prediction jointly contribute to accurate, stable, and view-consistent native 3D editing.

\paragraph{Operational Efficiency and Robustness.}
We further analyze the practical utility of VGGT-Edit in terms of efficiency and operational robustness. Unlike 2D-lifting methods whose latency increases with the number of views, VGGT-Edit maintains a constant, low-latency inference of approximately 2 seconds per scene due to its native feed-forward nature. This capability stems from our 3D residual learning paradigm, which empowers the model to map semantic intent to arbitrary point-level displacements while maintaining structural integrity. 

\subsection{Generalization to Unseen Instructions}

\begin{figure}[htbp]
  \centering
  \vspace{-0.2cm}
  \centerline{\includegraphics[width=0.9\linewidth]{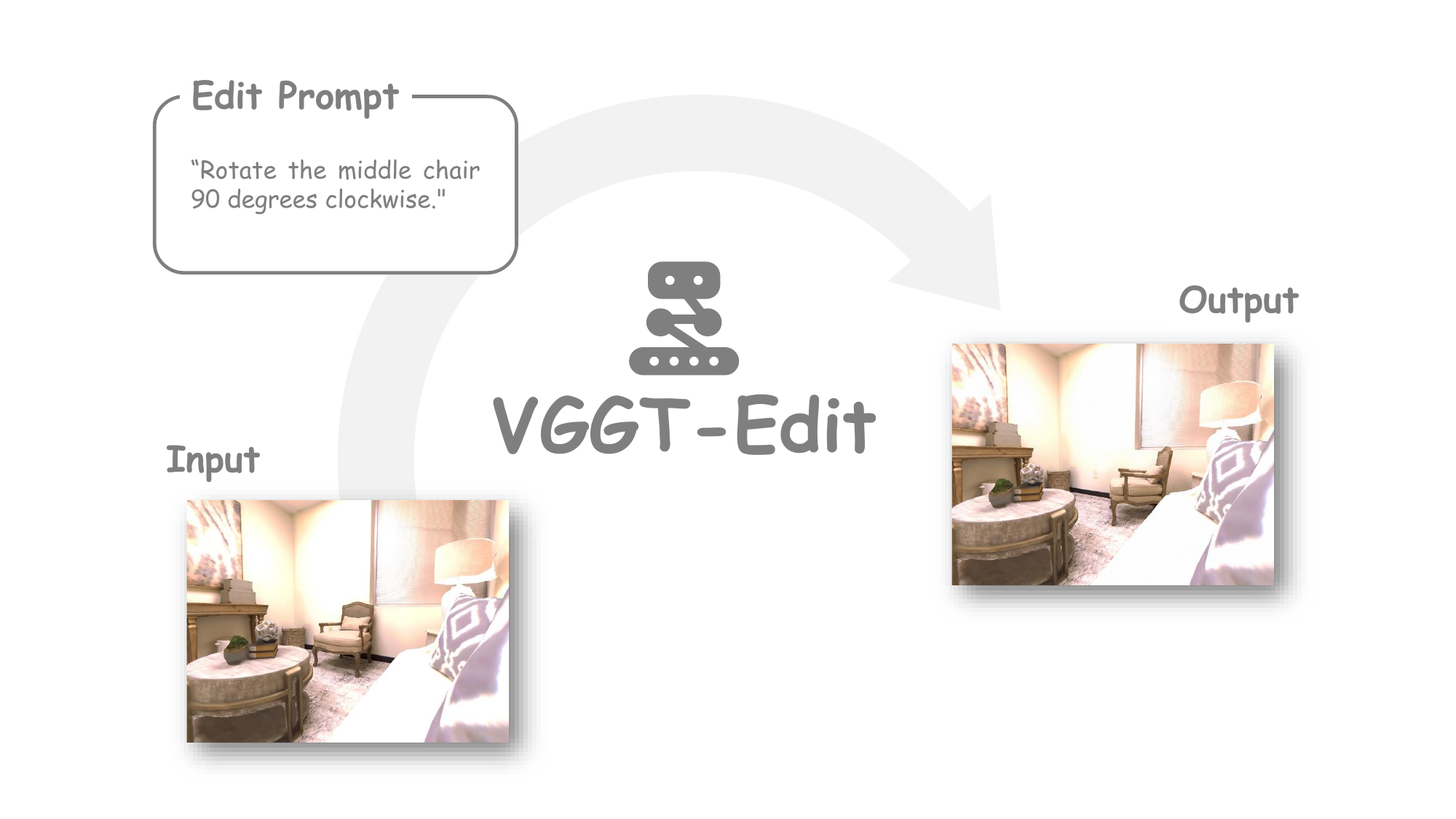}}
  \vspace{-0.2cm}
  \caption{Generalization to Unseen Instructions.}
  \label{fig:Unseen}
\end{figure}

Beyond the primary editing operations used during training, such as addition, removal, relocation, and attribute transformation, we observe that VGGT-Edit demonstrates remarkable generalization to unseen text instructions. This zero-shot capability allows the model to execute complex geometric commands that were not explicitly included in the training set. For instance, as shown in Fig \ref{fig:Unseen}, when given an instruction like "rotate the middle chair 90 degrees clockwise." our model can successfully produce the corresponding geometric shift while maintaining the chair's structural integrity. This flexibility stems from our 3D residual field prediction paradigm. Instead of simply memorizing predefined actions, the model learns a fundamental mapping between semantic intent and raw point-level displacements ($\boldsymbol{\Delta}\mathbf{P}$). By understanding how to translate text guidance into precise spatial deformations, VGGT-Edit can synthesize arbitrary motion fields to satisfy diverse human requests. This confirms that our framework has captured a deep, generalizable understanding of 3D spatial manipulation, making it highly effective for real-world interactive applications.

\begin{figure}[ht]
    \centering
    \includegraphics[width=0.92\linewidth]{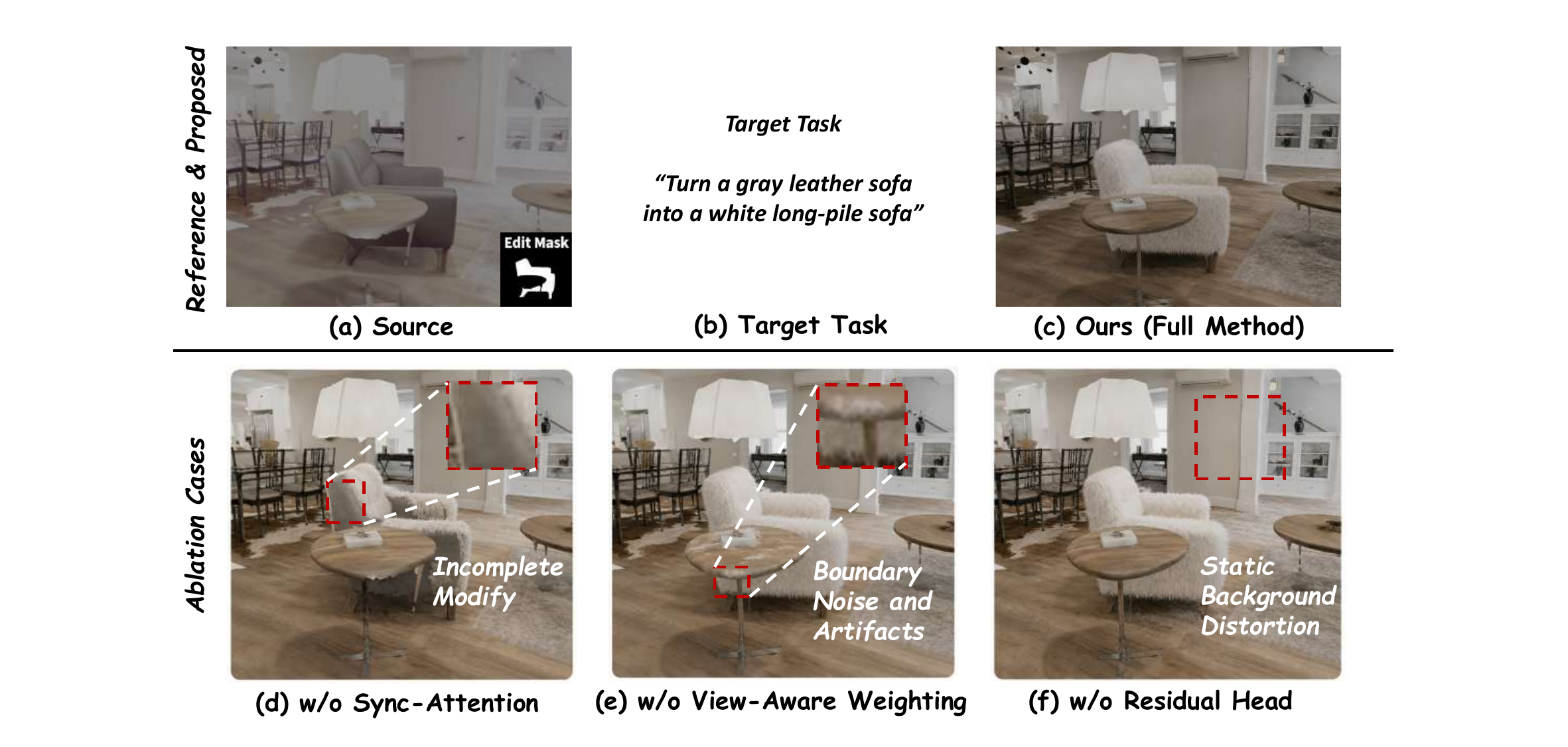}
    \caption{\textbf{Qualitative ablation study.} 
    Given the source scene (a) and the editing task (b), our full method (c) produces a coherent result. 
    Removing Sync-Attention (d) leads to incomplete material modification; 
    removing View-Aware Weighting (e) introduces boundary noise and artifacts near occluded regions; 
    removing the Residual Head (f) causes subtle distortion in static background areas.}
    \label{fig:ablation_study}
\end{figure}

\subsection{Qualitative Analysis}
To further illustrate the impact of each component in VGGT-Edit, we provide qualitative comparisons in Figure~\ref{fig:ablation_study}. The visual evidence aligns with our quantitative findings:

\textbf{w/o Depth-Synchronized Attention:} Without rhythmic semantic reinforcement, the model exhibits "incomplete editing" or color mismatches, failing to fully manifest the requested changes (e.g., the added object appears faded or incorrectly textured).

\textbf{w/o View-Aware Weighting:} The absence of this module leads to noticeable geometric artifacts and "floaters," particularly at the edges of occluded regions where the model fails to resolve spatial ambiguities.

\textbf{w/o Residual Transformation Head:} Replacing our residual paradigm with a full reconstruction head causes significant background "drifting." Background objects that should remain static exhibit subtle deformations or shifts, confirming the importance of our 3D residual learning for maintaining structural integrity.

\section{Conclusion}
In this paper, we present VGGT-Edit, a feed-forward framework for instruction-driven native 3D scene editing. By treating editing as a 3D residual field prediction task, our model avoids the multi-view inconsistencies common in 2D-lifting pipelines while preserving the structural integrity of the original scene. Through depth-synchronized text injection and view-aware weighting, VGGT-Edit achieves precise semantic-to-spatial alignment and robust feature fusion, enabling localized geometry deformations in a single forward pass. Experimental results on the DeltaScene Dataset show that our method outperforms existing optimization-based and feed-forward baselines in terms of geometric fidelity, multi-view consistency, and inference efficiency. With its ability to handle diverse operations and maintain high-speed performance, VGGT-Edit provides a practical and generalizable solution for real-time, interactive 3D scene editing.




{
\small
\bibliographystyle{plainnat}
\bibliography{main}





\end{document}